\documentclass{article}
\usepackage{spconf,graphicx}
\usepackage{amsmath}
\usepackage{booktabs}

\usepackage{amssymb}

\title{Using Conditional Generative Adversarial Networks to Generate Ground-Level Views From Overhead Imagery}

\name{Xueqing Deng, Yi Zhu and Shawn Newsam}
\address{University of California, Merced}
\begin{document}
%\ninept
%
\maketitle
\begin{abstract}
This paper develops a deep-learning framework to synthesize a ground-level view of a location given an overhead image. We propose a novel conditional generative adversarial network (cGAN) in which the trained generator generates realistic looking and representative ground-level images using overhead imagery as auxiliary information. The generator is an encoder-decoder network which allows us to compare low- and high-level features as well as their concatenation for encoding the overhead imagery. We also demonstrate how our framework can be used to perform land cover classification by modifying the trained cGAN to extract features from overhead imagery. This is interesting because, although we are using this modified cGAN as a feature extractor for overhead imagery, it incorporates knowledge of how locations look from the ground.
\end{abstract}
\section{Introduction}
\label{sec:intro}
% \\\texttt{icip2019@cmsworkshops.com}.

Mapping geographic phenomena on the surface of the Earth is an important scientific problem. The widespread availability of geotagged social media has enabled novel approaches to geographic discovery. In particular, \emph{proximate sensing}~\cite{leung2010proximate,Newsam2019journal}, which uses ground-level images and videos available at sharing sites like Flickr and YouTube, provides a different perspective from remote sensing, one that can see inside buildings and detect phenomena not observable from above. Proximate sensing has been applied to map land use~\cite{zhu2019tmm,zhu2015land}, public sentiment~\cite{zhu2016spatio}, human activity~\cite{Zhu2017Activity}, air pollution~\cite{li2015using}, and natural events~\cite{wang2016tracking}, among other things. However, a fundamental limitation to using geotagged social media for mapping is its sparse and uneven spatial distribution. Unlike overhead imagery, it generally does not provide dense or uniform coverage~\cite{deng2018icip}--it is not available everywhere.

\begin{figure}
    \centering
    \includegraphics[width=\linewidth]{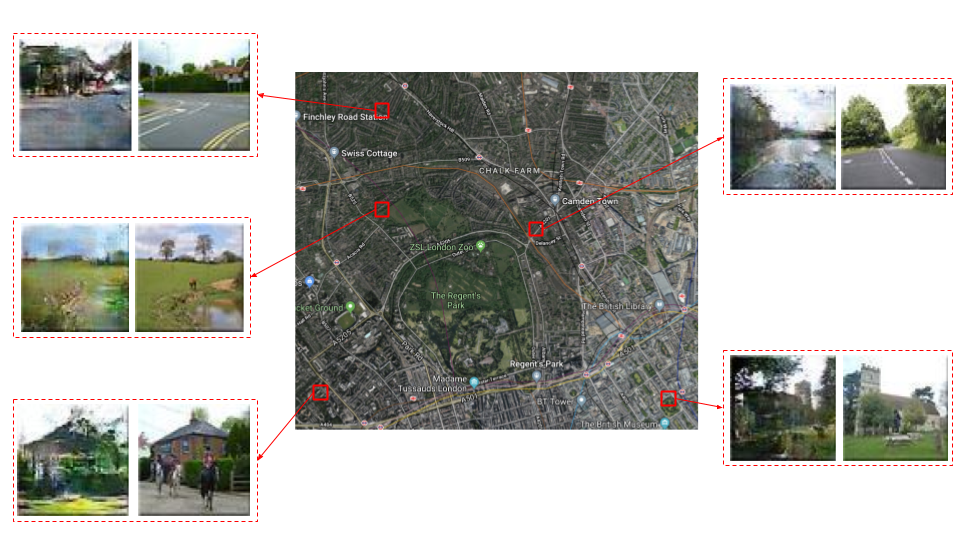}
    \caption{We develop a novel cGAN framework that takes an overhead image as input and generates ground-level views. Pictured above are pairs of ground-level images synthesized using our method (left) and real images (right) for various locations.}
    \label{fig:overview}
\end{figure}

In this paper, we try to get around this limitation of ground-level images by trying to answer the question ``what does this location look like at ground view based on overhead imagery?'' That is, we synthesize ground-level images using overhead imagery. Since overhead imagery is available everywhere, we can generate ground-level images everywhere. Figure~\ref{fig:overview} shows sample ground-level images generated using our framework.

We use a conditional generative adversarial network (cGAN) to generate ground-level images using the overhead imagery as auxiliary information. In previous work~\cite{deng2018sigspatial}, we used simple embeddings of the overhead imagery such as the grayscale values in a small image patch to condition the cGAN. In this paper, we develop a novel, more powerful network that allows us to condition the cGAN on significantly more of the overhead image. We do this by designing the generator of the cGAN as an encoder-decoder network in which the encoder feature maps are used as the auxiliary information.

Cross-view image synthesis has been investigated but only using Google Street View images~\cite{Regmi_2018_CVPR,2018arXivcrossview,zhai2017cvpr}. These approaches are limited in that they can only be used to generate ground-level images at street locations and they are arguably solving an easier problem since there is less scene diversity. Researchers have also investigated methods to get around the sparsity of ground-level images by interpolating features extracted from the images but this again was only using Street View images and, as we have shown in previous work~\cite{deng2018icip}, such interpolation methods are problematic since they fail to infer the spatially discontinuous nature of the surface of the Earth.

The contributions of our work are as follows:
(1) We propose a novel cGAN based on an encoder-decoder network for generating ground-level views given overhead imagery. This network allows the cGAN to be conditioned on more of the overhead imagery.
(2) We explore different levels of feature maps in the encoder as well as their concatenation to encode the overhead imagery.
(3) We demonstrate that our framework can be used to perform land cover classification.

%Our paper is organized as follows. Section 2 provides the technical details of our framework. Section 3 describes the datasets, and provides visualizations of the generated ground-level views, and presents quantitative results of the image synthesis and land-cover classification. Section 5 concludes.

\section{Methodology}
\label{sec:method}

\begin{figure}
    \centering
    \includegraphics[width=\linewidth]{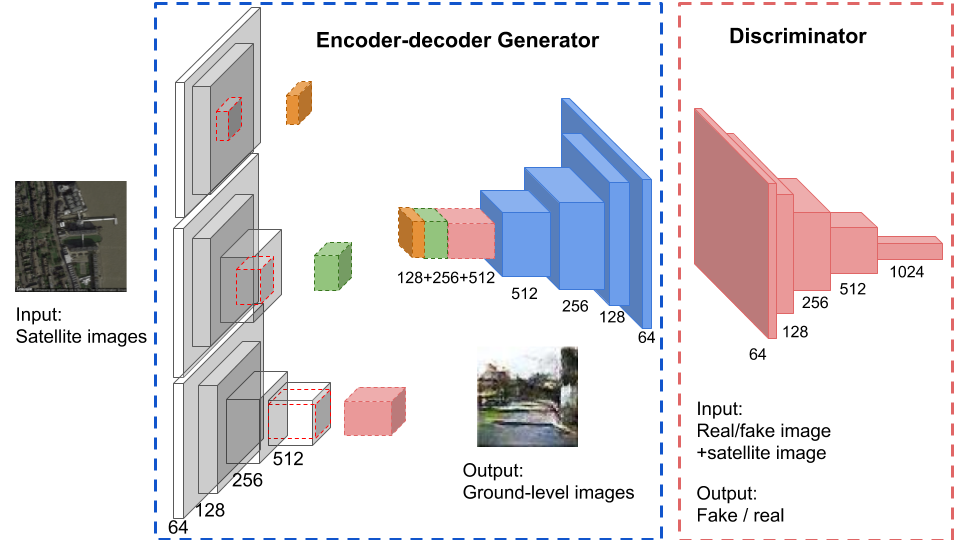}
    \caption{The framework of our proposed method. The generator takes as input an overhead image and generates a ground-level image. The discriminator tries to distinguish between the real and generated ground-level images. The generator has an encoder-decoder structure. The cGAN is conditioned using different feature map layers of the encoder as well as their concatenation. See the text for more details.}
    \label{fig:framework}
\end{figure}

\subsection{Background on GANs}
GANs \cite{goodfellow2014generative} consist of two components, a generator and a discriminator. In vanilla GANs, the generator $G$ generates realistic looking but ``fake'' images by upsampling random noise. The discriminator's goal is to distinguish between real and fake images. GANs learn generative models through adversarial training. That is, $G$ tries create images that look so realistic that $D$ cannot tell they are fake. $G$ and $D$ are trained simultaneously. $G$ is optimized to learn the true data distribution by generating images that are difficult to distinguish from real images by $D$. Meanwhile, $D$ is optimized to differentiate fake images generated by $G$ from real images. Overall, the training procedure is similar to a two-player min-max game with the following objective function

\begin{equation}
\begin{split}
\min \limits_{G}\max \limits_{D} V(D,G)=\mathbb{E}_{x\sim p_{data}(x)}[\log D(x)] + \\
\mathbb{E}_{z\sim p_{data}(z)}[1-  \log D(G(z))]
\end{split}
\end{equation}

\noindent where $z$ is the random noise vector and $x$ is the real image. Once trained, $G$ can be used to generate new, unseen images given random vectors as input.

% In practice, rather than train $G$ to minimize $1- \log D(G(z))$, we train $G$ to maximize $\log D(G(z))$, as demonstrated in \cite{goodfellow2014generative}.
%While training, we alternate between optimizing $D$ and $G$ for each epoch. 

Vanilla GANs learn the distribution of the entire training dataset. Conditional GANs (cGANs) were subsequently introduced to learn distributions conditioned on some auxiliary information. For example, a GAN can be trained to generate realistic-looking images of hand written digits. However, if we want images of a specific digit, a ``1'' for example, the generative model needs to be conditioned on this information. In cGANs, the auxiliary information $y$ is incorporated through hidden layers separately in both the generator and discriminator. $y$ can take many forms, such as class labels \cite{mirza2014conditional}. The cGAN objective function becomes

\begin{equation}
\begin{split}
\min \limits_{G}\max \limits_{D} V(D,G)=\mathbb{E}_{x\sim p_{data}(x)}[\log D(x| y)] + \\
\mathbb{E}_{z\sim p_{data}(z)}[1-  \log D(G(z|y))].
\end{split}
\end{equation}

\subsection{Proposed cGAN-based Framework}

Our goal is to generate ground-level views of a location given overhead imagery. The overhead imagery thus serves as our auxiliary information. A key question is, however, how to best incorporate this auxiliary information into the cGAN.

In our previous work~\cite{deng2018sigspatial}, we investigated different \emph{embeddings} to encode the raw pixel information from the overhead imagery. It is these embeddings that were input to the cGAN. We considered the color of the pixels in a 10$\times$10 patch of the overhead imagery, the grayscale values of this patch, and features extracted using a pretrained CNN model.

In this paper, we instead propose a cGAN in which the generator is an encoder-decoder network in which the feature maps from the encoding stage serve as the auxiliary information. This addresses two shortcomings of our previous approach. First, the cGAN can ``see'' more than a 10$\times$10 pixel patch in the overhead image when generating a ground-level image. Second, we learn the optimal feature embedding of the overhead image data.

Our framework is shown in figure~\ref{fig:framework}. This figure actually shows three variants of our network which differ in the number of convolutional layers in the encoder. In one variant, corresponding to the top portion of the generator subfigure, the encoder consists of just two convolutional layers. We crop a $4\times4$ region (red dashed region) from the feature maps after the second convolutional layer and pass this to the decoder part of the generator. In this variant, this $4\times4$ region can be thought of as low-level features that are derived from a smaller receptive field of the overhead image.

In the variant corresponding to the bottom portion of the generator subfigure, the encoder has four convolutional layers and the $4\times4$ region is extracted from the feature maps after the fourth convolutional layer. This $4\times4$ region can be thought of as high-level features that are derived from a larger receptive field of the overhead image. The final variant contains three convolutional layers and extracts mid-level features.

These three variants allow us to consider whether low-level features or high-level features are better for generating the ground-level views. They also allow us to investigate the effects of different receptive field sizes.

We consider three cases in the experiments below: 1) just using the low-level features (orange cube in figure~\ref{fig:framework}) ; 2) just using the high-level features (pink cube); and 3) ~\emph{concatenating} the low-, middle-, and high-level features (orange-green-blue cube).

\subsection{Network Details}

The generator is an encoder-decoder network. It takes as input a 128$\times$128 pixel overhead image and outputs a synthesized 64$\times$64 pixel ground-level view. The encoder and decoder components consist of convolution layers, batch normalization~\cite{ioffe2016batchnorm}, and activation layers. Leaky ReLU with a slope of 0.2 is used as the activation function in the encoder whereas ReLU is used in the decoder except for the final layer where tanh is used. Strided convolutions are used to increase or decrease the feature map resolution instead of maxpooling.

The discriminator is similar to the encoder of the generator but without feature map cropping.  It follows the same conv-batchnorm-leakyReLU architecture but its last layer is a sigmoid activation function that outputs a binary value corresponding to real or fake. The discriminator takes as input a ground-level image, real or fake, and the auxiliary information in the form of the overhead image. It outputs a binary value indicating whether it thinks the input ground-level image is real or fake. In the case of a real image, the overhead image is the actual overhead view of where the real image is located. In the case of a fake image, the overhead image is what was used by the generator to produce the image. The output of the discriminator is its belief of whether the ground-level image is real or fake. Two different losses are used to train the discriminator depending on whether the input ground-level image is real or fake.

Our proposed cGAN architecture has the following objective functions

\begin{equation}
\begin{split}
&L_D=\mathbb{E}_{(I_g,\varphi(I_o)) \sim p_{data}}(I_g,I_o)+ \\
&\mathbb{E}_{\varphi(I_o)\sim p_{data}}[1-  \log D(G(\varphi(I_o)),I_o)]
\end{split}
\end{equation}

\begin{equation}
L_G=\mathbb{E}_{\varphi(I_o)\sim p_{data}}[1-  \log D(G(\varphi(I_o)),I_o)]
\end{equation}
where $I_o$ and $I_g$ refer to overhead and ground-level images respectively, and $\varphi(I)$ is the cropped feature maps or their concatenation.

As mentioned earlier, our goal is not just to create realistic looking ground-level views from overhead imagery but learn a representation that is useful for other tasks, specifically land-cover classification. Towards this end, following~\cite{radford2015unsupervised}, we convert our trained cGAN into a feature extractor by removing the last sigmoid layer of the discriminator and adding an average pooling layer to generate a 1024D feature vector. Given an overhead image of a location, we can now extract a feature vector that can be used to classify that location. Importantly, this feature extractor has been learned \emph{with knowledge of what this location looks like from ground level}.

\begin{figure}
\includegraphics[width=\linewidth]{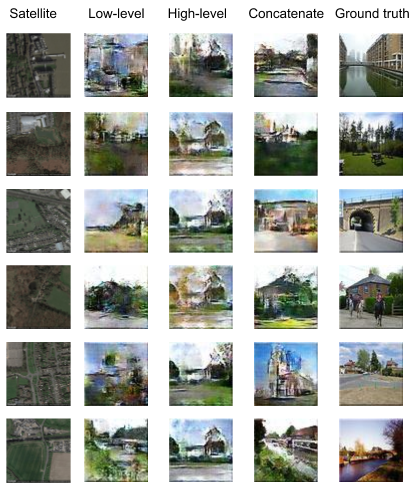}
%\vspace{-2mm}
\caption{Examples of ground-level views generated with our framework. The leftmost column shows the overhead images used to generate the views. Columns two through four show images generated with low-, high-, and concatenated features. The rightmost column shows the real ground-level view at the location of the overhead image.}
\label{fig:result}
\end{figure}

\section{Experiment One: Generating Ground-Level Views}
\label{sec:experiment}
This first experiment investigates the ground-level images generated by our framework. Specifically, given an overhead image, we use the generator component of our trained cGAN network to generate a ground-level view. We then compare this with the real ground-level view.

\subsection{Dataset and Training}
We need co-located ground-level and overhead imagery to train our cGAN network. We download ground-level images with known location using the Geograph API\footnote{http://www.geograph.org.uk/}. We download co-located georeferenced overhead imagery using the Google Map Static API\footnote{https://developers.google.com/maps/documentation/maps-static/intro}.

We train our cGAN using 4,000 co-located Geograph and overhead image pairs.  We implement our deep learning framework using the PyTorch\footnote{https://pytorch.org} framework and Adam \cite{Kingma2014AdamAM} is used as our optimizer. The initial learning rate is set to 0.0002. We train our cGAN for 400 epochs with a batch size of 128 on one NVIDIA TITAN Xp GPU.

\subsection{Qualitative results}
We generate three sets of ground-level views given overhead images: 1) views generated with the low-level features from the encoder; 2) views generated with the high-level features; and 3) views generated with the concatenation of the low-, middle-, and high-level features. Figure~\ref{fig:result} shows these views for different overhead images along with the real ground-level views which we consider as the ground truth. These results show that the framework is able to learn the transformation between views. In particular, the ground-level images generated
using the concatenation of different feature levels are detailed with regions corresponding to roads, trees,
sky, grass, and houses. Their similarity with the ground truth shows they do characterize what the locations look like from the ground. The images generated using the low-level features are not as detailed and not as visually similar to the ground truth. The images generated using the high-level features are visually alike, indicating mode collapse during training. (We comment on this further below.)

%For land-cover classification, we use the ground-truth land-cover map LCM2015\footnote{https://eip.ceh.ac.uk/lcm/lcmdata} to construct our training and test datasets. This map provides land-cover classes for the entire United Kingdom on a 1km grid. We group these classes into two super-classes, urban and rural, and limit our study to a 71km$\times$71km region containing London. The Geograph images are labeled as urban or rural based on the LCM2015 label of the 1km$\times$1km grid cell from which they are downloaded. We realize this label propagation likely results in some noisy labels in our dataset. 

\subsection{Quantitative Results}
A common quantitative GAN evaluation measure for image synthesis is the inception score \cite{2016nipsImproved}. The core idea behind the inception score is to assess how diverse the generated samples are within a class while being meaningfully representative of the class. We compute an inception score of our generated images using an AlexNet~\cite{alexnet2012nips} model trained on the Places dataset~\cite{zhou2018pami} since our outdoor images are aligned with the classes in this data. A higher inception score indicates the set of generated images is more diverse (a desired result). Table~\ref{tab:quantitative} shows that, based on this metric, the concatenation of different level feature levels is also quantitatively the best method.

\section{Experiment Two: Land Cover Classification}
In this experiment, we use our cGAN framework to classify overhead imagery into rural and urban locations. Specifically, we use our framework to extract features from overhead imagery which are then classified using an SVM classifier.

\subsection{Dataset and Training}
We select 20,000 locations and label them as rural or urban using the ground-truth land cover map LCM2015\footnote{https://eip.ceh.ac.uk/lcm/lcmdata}. We extract a 1024D feature from each of these locations using overhead imagery. 4,000 of these features are used to train a binary SVM classifier. We evaluate this classifier using the remaining 16,000 features.

\subsubsection{Classification Accuracy}
Table~\ref{tab:quantitative} lists the land cover classification accuracy of our cGAN network using the low-level, high-level, and concatenated features. All three results are seen to be better than our previous method~\cite{deng2018sigspatial} which used the grayscale values from a 10$\times$10 pixel patch in the overhead image as the auxiliary information for the cGAN. The proposed encoder-decoder generator is thus seen to be an improvement. While we showed above that the concatenated features produce more realistic looking ground-level images, the high-level features are seen to perform better for classification. This is reasonable since classification is a high-level task. The high-level features are good for high-level tasks but do not contain enough detail information for image synthesis and thus the mode collapse that was observed above.

\begin{table}[htbp]
\caption{Quantitative results of image synthesis and land-cover classification.}
\label{tab:quantitative}
\centering
\begin{tabular}{ p{3cm}|p{2cm} |p{2cm} }
\toprule
Methods&Inception Score& Classification Accuracy (\%)\\
\hline
Previous method~\cite{deng2018sigspatial} & - & 82.3\\
Low-level&1.945& 83.58\\
High-level& 1.241 & \bf{86.71}\\
Concatenate& \bf{2.526}  &  85.45\\
\bottomrule
\end{tabular}
\end{table}

\section{Conclusion}
\label{sec:conclusion}
We propose a novel cGAN framework for generating ground-level views given overhead imagery. By using an encoder-decoder network as the generator, we can condition the cGAN on more of the overhead imagery than a previous approach which uses simple embeddings. We compare conditioning the cGAN on different feature map levels from the encoder as well their concatenation. We show the approach generates realistic and representative ground-level images. We also show that the proposed framework can be used for land cover classification and outperforms our previous approach on this task.

%We explored image generation using conditional GANs with a encoder-decoder generator for generating ground-level views and the corresponding image features given overhead imagery. The generated ground-level images look more natural than our previous work. Although, we still suffer the same problem of the lack the details of real images. The concatenation of feature maps from different intermediate layers improve the image quality, but harness the classification accuracy which will be studied in the future. However, they do capture the visual distinction between urban and rural scenes and improve the classification. 

%\section{Acknowledgements}
%This work was funded in part by a National Science Foundation CAREER grant, \#IIS-1150115. We gratefully acknowledge the support of NVIDIA Corporation through the donation of the GPU card used in this work.

\clearpage

% -------------------------------------------------------------------------
\bibliographystyle{IEEEbib}
\bibliography{refs}

\begin{thebibliography}{10}

\bibitem{leung2010proximate}
D.~Leung and S.~Newsam,
\newblock ``Proximate sensing: Inferring what-is-where from georeferenced photo
  collections,''
\newblock in {\em CVPR}, 2010.

\bibitem{Newsam2019journal}
S.~Newsam and D.~Leung,
\newblock {\em Georeferenced Social Multimedia as Volunteered Geographic
  Information}, pp. 225--246,
\newblock 2019.

\bibitem{zhu2019tmm}
Y.~Zhu, X.~Deng, and S.~Newsam,
\newblock ``Fine-grained land use classification at the city scale using
  ground-level images,''
\newblock {\em IEEE Transactions on Multimedia}, 2019.

\bibitem{zhu2015land}
Y.~Zhu and S.~Newsam,
\newblock ``Land use classification using convolutional neural networks applied
  to ground-level images,''
\newblock in {\em ACM SIGSPATIAL}, 2015.

\bibitem{zhu2016spatio}
Y.~Zhu and S.~Newsam,
\newblock ``Spatio-temporal sentiment hotspot detection using geotagged
  photos,''
\newblock in {\em ACM SIGSPATIAL}, 2016.

\bibitem{Zhu2017Activity}
Y.~Zhu, S.~Liu, and S.~Newsam,
\newblock ``Large-scale mapping of human activity using geo-tagged videos,''
\newblock in {\em ACM SIGSPATIAL}, 2017.

\bibitem{li2015using}
Y.~Li, J.~Huang, and J.~Luo,
\newblock ``Using user generated online photos to estimate and monitor air
  pollution in major cities,''
\newblock in {\em ICIMCS}, 2015.

\bibitem{wang2016tracking}
J.~Wang, M.~Korayem, S.~Blanco, and D.~Crandall,
\newblock ``Tracking natural events through social media and computer vision,''
\newblock in {\em ACM MM}, 2016.

\bibitem{deng2018icip}
X.~Deng, Y.~Zhu, and S.~Newsam,
\newblock ``Spatial morphing kernel regression for feature interpolation,''
\newblock in {\em ICIP}, 2018.

\bibitem{deng2018sigspatial}
X.~Deng, Y.~Zhu, and S.~Newsam,
\newblock ``What is it like down there?: Generating dense ground-level views
  and image features from overhead imagery using conditional generative
  adversarial networks,''
\newblock in {\em ACM SIGSPATIAL}, 2018.

\bibitem{Regmi_2018_CVPR}
K.~Regmi and A.~Borji,
\newblock ``Cross-view image synthesis using conditional gans,''
\newblock in {\em CVPR}, 2018.

\bibitem{2018arXivcrossview}
K.~Regmi and A.~Borji,
\newblock ``{Cross-view image synthesis using geometry-guided conditional
  GANs},''
\newblock {\em ArXiv e-prints}, 2018.

\bibitem{zhai2017cvpr}
M.~Zhai, Z.~Bessinger, S.~Workman, and N.~Jacobs,
\newblock ``Predicting ground-level scene layout from aerial imagery,''
\newblock in {\em CVPR}, 2017.

\bibitem{goodfellow2014generative}
I.~Goodfellow, J.~Pouget-Abadie, M.~Mirza, B.~Xu, D.~Warde-Farley, S.~Ozair,
  A.~Courville, and Y.~Bengio,
\newblock ``Generative adversarial nets,''
\newblock in {\em NIPS}, 2014.

\bibitem{mirza2014conditional}
M.~Mirza and S.~Osindero,
\newblock ``Conditional generative adversarial nets,''
\newblock {\em arXiv preprint}, 2014.

\bibitem{ioffe2016batchnorm}
S.~Ioffe and C.~Szegedy,
\newblock ``Batch normalization: Accelerating deep network training by reducing
  internal covariate shift,''
\newblock in {\em ICML}, 2015.

\bibitem{radford2015unsupervised}
A.~Radford, L.~Metz, and S.~Chintala,
\newblock ``Unsupervised representation learning with deep convolutional
  generative adversarial networks,''
\newblock in {\em ICRL}, 2016.

\bibitem{Kingma2014AdamAM}
D.~P. Kingma and J.~Ba,
\newblock ``Adam: A method for stochastic optimization,''
\newblock {\em CoRR}, 2014.

\bibitem{2016nipsImproved}
T.~Salimans, I.~Goodfellow, W.~Zaremba, V.~Cheung, A.~Radford, and X.~Chen,
\newblock ``Improved techniques for training gans,''
\newblock in {\em NIPS}, 2016.

\bibitem{alexnet2012nips}
A.~Krizhevsky, I.~Sutskever, and G.~Hinton,
\newblock ``Imagenet classification with deep convolutional neural networks,''
\newblock in {\em NIPS}, 2012.

\bibitem{zhou2018pami}
B.~Zhou, A.~Lapedriza, A.~Khosla, Au. Oliva, and A.~Torralba,
\newblock ``Places: A 10 million image database for scene recognition,''
\newblock {\em IEEE transactions on pattern analysis and machine intelligence},
  pp. 1452--1464, 2018.

\end{thebibliography}

\end{document}